# CircFormerMoE: An End-to-End Deep Learning Framework for Circular RNA Splice Site Detection and Pairing in Plant Genomes

Tianyou Jiang

Faculty of Medicine, University of Bern, Bern 3012, Switzerland
e-mail: tianyou.jiang.research@gmail.com

*Abstract*—Circular RNAs (circRNAs) are important components of the non-coding RNA regulatory network. Previous circRNA identification primarily relies on high-throughput RNA sequencing (RNA-seq) data combined with alignment-based algorithms that detect back-splicing signals. However, these methods face several limitations: they can't predict circRNAs directly from genomic DNA sequences and relies heavily on RNA experimental data; they involve high computational costs due to complex alignment and filtering steps; and they are inefficient for large-scale or genome-wide circRNA prediction. The challenge is even greater in plants, where plant circRNA splice sites often lack the canonical GT-AG motif seen in human mRNA splicing, and no efficient deep learning model with strong generalization capability currently exists. Furthermore, the number of currently identified plant circRNAs is likely far lower than their true abundance. In this paper, we propose a deep learning framework named CircFormerMoE based on transformers and mixture-of-experts for predicting circRNAs directly from plant genomic DNA. Our framework consists of two subtasks known as splicing site detection (SSD) and splicing site pairing (SSP). The models' effectiveness has been validated on gene data of 10 plant species. Trained on known circRNA instances, it is also capable of discovering previously unannotated circRNAs. In addition, we performed interpretability analyses on the trained model to investigate the sequence patterns contributing to its predictions. Our framework provides a fast and accurate computational method and tool for large-scale circRNA discovery in plants, laying a foundation for future research in plant functional genomics and non-coding RNA annotation

## I. INTRODUCTION

Circular RNAs (circRNAs) are important components of the non-coding RNA regulatory network [1], [2]. Owing to their stable, closed-loop structure, circRNAs can act as molecular sponges for microRNAs (miRNAs), thus modulating gene expression pathways [3]. They are also increasingly recognized as gene expression regulators, and biomarkers for diseases such as cancer and neurological disorders [4], [5]. These molecules interact with RNA-binding proteins (RBPs), modulate cellular signaling, and are notable for their abundance and conservation across species [6], [7]. In plants, circRNAs also emerge as potential regulators of development, stress response, and gene expression, presenting new opportunities for molecular breeding and crop improvement [10]. However, detecting and identifying plant circRNAs remains challenging due to their typically low abundance, tissue specificity, and the absence of canonical splicing signals, such as the over 95% GT-AG motif usage at 5′ and 3′ splice sites observed in human precursor mRNAs [11]. Compared to human mRNAs, plant circRNAs lack such well-defined splice signals and still lack effective computational tools. As a result, the number of known plants circRNAs is likely to be significantly underestimated [12], [13].

Predicting circRNAs directly from plant gene sequences is both biologically relevant and computationally challenging [14]. Accurate identification of circRNAs and their splicing patterns offers important insights into gene regulation mechanisms and the circularization process. From a computational perspective, circRNA prediction is a sequence-to-site or sequence-to-region task that often requires modeling long-range dependencies. This is because circularization typically involves detecting and pairing ultra distant splice sites. Traditional detection tools such as CIRCexplorer2 [8] and find_circ [9] rely on high-throughput RNA sequencing (RNA-seq) combined with alignment-based algorithms that search for back-splice junctions. Although effective in animals, these approaches face several limitations: they rely heavily on experimental RNA data, require computationally intensive alignment and filtering procedures. More importantly, these tools are not designed to predict circRNAs directly from plant genomic DNA, making them unsuitable for large-scale or genome-wide screening in plant genomes. Addressing these challenges with deep learning requires advanced modeling frameworks, such as that based on modules of transformers, attention-based recurrent networks, and graph neural networks.

Recent advances in deep learning have demonstrated the feasibility of predicting splice junctions from nucleotide sequences of human, as shown by models such as SpliceAI [15]. However, predicting plant circRNAs poses extra challenges. Plant genomes often exhibit long-range and non-canonical splicing patterns, with circRNA lengths can range from fewer than 50 to over 100,000 nucleotides. These characteristics challenge the capabilities of conventional deep learning models such as SpliceAI, which were originally designed for predicting splice junctions in human pre-mRNA sequences. Existing studies applying deep learning models for plant circular RNA prediction have primarily focused on processing short-read sequences, such as sequence-level classification approaches [16], [17], [18]. However, these classification models often fall short in practical applications due to the ultra-long nature of complete gonomes. For instance, plant genomes can span tens of millions to several billion base pairs, such short sequence level classification tasks are insufficient to handle such genomes.

In this paper, we address the task of circRNA prediction from plant genomic DNA sequence by introducing a novel deep neural network, CircFormerMoE (Circular RNA Splicing Sites Location Transformer with Mixture of Experts). Our exprimental data includes genome from 10 plant species: Brachypodium distachyon, Brassica juncea, Brassica napus, Cajanus cajan, Glycine max, Gossypium raimondii, Lactuca sativa, Medicago truncatula, Pisum sativum, and Theobroma cacao. The deep learning model is designed to predict circRNAs directly from plant genomic DNA sequences and composed of two subtasks: (a) Splicing Site Detection (SSD), a sequence-to-sequence task where the model detects splicing sites from inputs, and (b) Splicing Site Pairing (SSP), a sequence-to-class task where the model predicts whether pairs of splicing sites form circular RNAs. The mixture-of-experts concept is implemented by distributing species-specific heads for different species with a shared backbone, and our experiments demonstrate that this approach achieves better performance compared to training individual models for each species. After training, CircFormerMoE identifies substantially more circular RNA splicing sites than previously reported, demonstrating its efficiency for circRNA discovery. Additionally, we provide preliminary interpretability analyses to gain insights into the model's decision-making. The code, data and trained models used in the experiments are released at https://github.com/SkyCol/CircFormerMoE.

## II. MATERIALS AND METHODS

### A. Data

To achieve comprehensive detection of circular RNAs (circRNAs) directly from raw DNA sequences of plants, we constructed datasets for two tasks seperately: splicing site detection (SSD) and splicing site pairing (SSP). We utilized annotation files containing circRNA splicing sites positions from the PlantCircRNA database [19] and genome data in FASTA format from EnsemblPlants [20]. Data of 10 plant species were included in this paper, including Brachypodium distachyon (purple false brome), Brassica juncea (brown mustard), Brassica napus (rapeseed), Cajanus cajan (pigeon pea), Glycine max (soybean), Gossypium raimondii (cotton), Lactuca sativa (lettuce), Medicago truncatula (barrel medic), Pisum sativum (pea), and Theobroma cacao (cacao). An overview of circular RNA data used in this study is shown in **Fig. 1**. The distance distribution of paired circRNA splicing sites, also known as circRNA length among these 10 species is as follows: short distance (<1000 bases): 106626 samples for 82%; Medium distance (1000-5000 base pairs): 8788 samples for 6.8%; Long distance (5000-10000 base pairs): 4001 samples for 3.1%; Ultra long distance (>10000 bases): 10563 samples for 8.1%. Detailed information of datasets used for training and evaluation is shown in Table I, where *N/P* Ratio represents the negative and positive ratio of the splicing position and none splicing position of the sequence used for training the task 1 splicing site detection, Pos represents positive samples and Neg represents negative samples in the task 2 splicing site pairing.

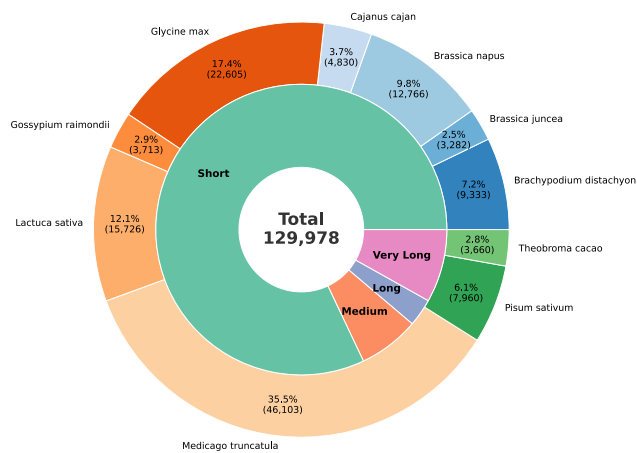

**Fig. 1.** Overview of circular RNA data used in this study. The outer circle shows the number of circRNAs obtained from 10 species. The inner circle shows the length distribution of all the included circRNAs.

TABLE I
DATASETS STRUCTURE.

| Gnome | Num | Task 1 | | Task 2 | |
|---|---|---|---|---|---|
| | | Samples | N/P | Pos | Neg |
| Brassica juncea | 3282 | 5952 | 873.34 | 3282 | 32699 |
| Brassica napus | 12766 | 22569 | 654.50 | 12766 | 107251 |
| B. distachyon | 9333 | 16316 | 581.76 | 9333 | 93330 |
| Cajanus cajan | 4830 | 7711 | 263.29 | 4830 | 41119 |
| Glycine max | 22605 | 34287 | 223.78 | 22605 | 225857 |
| Gossypium raimondii | 3713 | 6408 | 496.60 | 3713 | 37048 |
| Lactuca sativa | 15726 | 26515 | 368.68 | 26515 | 156700 |
| Medicago truncatula | 46103 | 63687 | 210.31 | 46103 | 459137 |
| Pisum sativum | 7960 | 12934 | 291.93 | 7960 | 75331 |
| Theobroma cacao | 3660 | 5340 | 227.54 | 3660 | 36531 |
| TOTAL | 129978 | 201719 | 285.35 | 129978 | 1265003 |

In Task 1 (SSD), the model is supported to detect candidate splicing sites from the genomic sequences. Scanned on each chromosome, a total of 201719 gene sequences with a length of 5001 were generated by truncating each splicing site and 2500 bases on both sides (missing values filled with "N"), along with corresponding labels. The label for each data point is annotated as a list, for example, [6, 2500, 2566] indicates that the 6th, 2500th and 2566th positions of this sample are splice sites of a circRNA. During training, every nucleotide position will get a binary label: 1 if the position corresponds to a circRNA splicing site (regardless of which circRNA it belongs to), and 0 otherwise. For example, the label "[6, 2500, 2566]" means the 6-th, 2500-th and 2566-th position will be set to 1 while others set to 0. Additionally, samples with no known splicing sites were also included by sampling from non-splicing regions of the genome with empty labels.

In Task 2 (SSP), the model is supported to pair each splicing sites detected by task 1 or currently known. When constructing the dataset, sequences of 1001 bases were extracted from each splicing site including 500 bases upstream and downstream. The sequences of the paired sites were concatenated to form positive samples, and unpaired site sequences were concatenated to form negative samples. To give a sense that the final samples were obtained by concatenating sequences of two sites, a short sequence of "MMMMM" is added in the middle of them to indicate the concatenation symbol. The ratio of positive and negative samples is set to approximately 1:10 and a total of 1394981 samples were constructed for this task.

### B. Model Framework

CircFormerMoE separates circRNA prediction into two tasks. Task 1, Splicing Sites Detection (SSD), involves detecting precise splicing site locations along raw DNA sequences in a position-wise manner. Task 2, Splicing Sites Pairing (SSP), determines which pairs of detected splicing sites form circular RNAs, classifying sequences accordingly, where input sequences are constructed by concatenating the genomic regions surrounding the candidate splice sites.

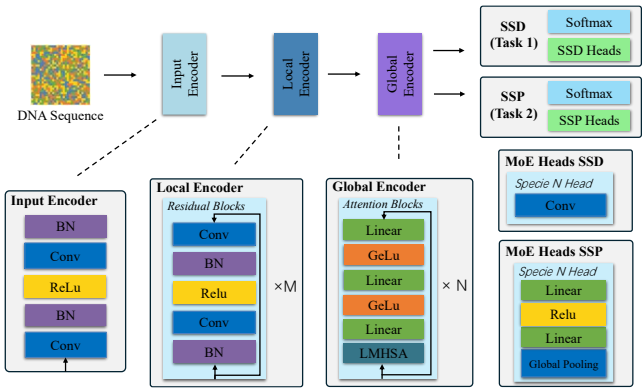

**Fig. 2.** CircFormerMoE Architecture. It consists of an input encoder, a local encoder, a global encoder and specific Mixture-of-Experts (MoE) heads for each species.

*1) Model Architecture Overview*

The overall architecture of CircFormerMoE is illustrated in Fig. 2. It consists of an input encoder, a local encoder, a global encoder and mixture-of-experts (MoE) heads different designed for Task 1 and Task 2. The input encoder transforms the one-hot encoded input sequence of five channels (A, C, G, T, N in Task 1 and A, C, G, T, N, M in Task 2) into a higher-dimensional representation. The local encoder, composed of multiple 1D convolutional residual blocks, captures local contextual features within the sequence. The global encoder further enhances representations by capturing long-range dependencies through several attention blocks based on a lightweight multi-head self-attention (LMHSA) design, which reduces computational cost while maintaining performance. Finally, the MoE heads consist of species-specific expert layers, where each head is assigned to each species to specialize in learning the splicing patterns characteristic of specific species.

*2) Mixture-of-Experts Heads*

CircFormerMoE employs Mixture-of-Experts (MoE) heads, where each expert is responsible for a specific species. The MoE design enables the model to leverage shared knowledge across species while allowing species-specific specialization through a small number of learnable parameters. All input sequences are processed through shared encoders. The shared backbone transforms the raw input into a high-dimensional representation $H \in \mathbb{R}^{L \times D}$, where $L$ is the sequence length and $D$ represent the hidden dimension.

During training, a hard routing mechanism is adopted: each sample is routed to its corresponding expert head $f_{s^*}$, where $s^*$ is the known species identity. Samples in each batch contain multiple species. This allows each expert to specialize in learning splicing characteristics unique to its species while benefiting from the common feature extractor.

Depending on the task, different operations are applied to the shared feature $H$:

$$f(x) = \begin{cases} f_{s^*}^{SSD}(H), & \text{for Task 1 (SSD)} \\ f_{s^*}^{SSP}(\text{Pool}(H)), & \text{for Task 2 (SSP)} \end{cases} \quad (1)$$

In Task 1 (SSD), the model outputs a position-wise prediction indicating the likelihood of each base being a splicing site. Each species-specific head $f_{s^*}^{SSD}$ is implemented as a 1×1 convolution layer that preserves sequence length and outputs a score for each base.

In Task 2 (SSP), the model performs a sequence-level classification to determine whether a pair of sites forms a circular RNA. The feature map $H$ is first aggregated using global average pooling, and the pooled vector is passed through a fully connected head $f_{s^*}^{SSP}$ to produce a single prediction score.

During inference, for Task 1, predictions over overlapping windows (e.g., window size 5001 with one-third overlapping) are averaged to smooth the output. For Task 2, sequences of candidate splice site pairs are concatenated, and the corresponding expert is applied to predict their category after the backbone.

*3) Lightweight Multi-Head Self-Attention*

The attention blocks in the global encoder adopt a lightweight multi-head self-attention (LMHSA) mechanism based on linear attention methods [21], [22]. In Task 1, the input sequence has a length of 5001. Using standard MHSA would result in a quadratic complexity of $O(5001^2)$, exceeding 25 million operations per layer, which is computationally expensive and impractical. Specifically, we employ the attention used in Performer, an improved linear attention that leverages random feature maps and the FAVOR+ algorithm [23]. The core idea is to approximate the attention:

$$\text{Attention}(Q, K, V) = \text{softmax}\left(\frac{QK^T}{\sqrt{d}}\right)V \quad (2)$$

With the following kernel-based formulation:

$$\text{Attention}(Q, K, V) = \phi(Q)(\phi(K)^T V) \quad (3)$$

Where $\phi$ is a random feature map that transforms queries and keys into a high-dimensional feature space, enabling the attention computation to scale linearly with the sequence length $O(n)$. A commonly used random feature map in Performer is:

$$\phi(x) = \frac{1}{\sqrt{m}} \exp\left(w^T x - \frac{\|x\|^2}{2}\right) \quad (4)$$

Where $w \sim \mathcal{N}(0, I)$ is drawn from a Gaussian distribution and $m$ is the number of random features controlling the approximation quality.

*C. Training and evaluation settings*

Due to class imbalance in the training data, before training, we computed the ratio of positive to negative sample positions in the training set, see Table 1. This ratio was then used to set a weight to the loss function, encouraging the model to focus more on underrepresented positive samples. Specifically, for the task of splicing site detection, we set the positive class weight to 285 for the first 10 epochs, reflecting the ratio of positive to negative positions. For the splicing site pairing task, a positive class weight of 10 was applied for the first 5 epochs, reflecting the ratio of positive to negative positions. Furthermore, training across all species introduced a double imbalance. To address this, we employed uniform sampling in each epoch, ensuring that an equal number of samples from each species were included in every batch.

CircFormerMoE was first pretrained on data from all general species. During this pretraining stage, it was compared against the baseline CircFormer model without species-specific expert heads. After 500 epochs of training on the splicing site detection task (Task 1), the pretrained CircFormerMoE was fine-tuned on each individual species for both Task 1 (splicing site detection) and Task 2 (splicing site pairing) to boost the performance. We then compared the resulting models with SpliceAI, as well as with our single-species-trained CircFormer model (without MoE) on Task 1.

During Task 1, we evaluated model performance using average precision (AP), average recall (AR), average f1-score, and top-k accuracy that introduced in SpliceAI to access the model's detection performance [15]. Here, "average" refers to the averaging across all sequence positions. For Task 2, which involves splicing site pairing, we adopted metrics including balanced accuracy, precision, recall and f1-score to assess the model's classification performance. It is important to note that the definition of top-k accuracy in this paper differs from the commonly used definition in image classification. The specific formulas for top-k accuracy and balanced accuracy used here are provided in equations (5) and (6).

$$Top-k = \frac{top\ k\ predicted\ pos\ \cap\ true\ splice\ pos}{k} \quad (5)$$

$$Balanced\ Accuracy = \frac{1}{2}\left(\frac{TP}{TP+FN} + \frac{TN}{TN+FP}\right) \quad (6)$$

Top-k accuracy is used because of the rarity of splice sites in task 1. For a given class, let $k$ is the number of true splice site positions in the test set. We identify the top-k predicted positions using a peak detection algorithm applied to the model's output probability map. The top-k accuracy is then defined as the proportion of true splice sites that are recovered within the top-k predicted peaks. This metric is equivalent to the precision at the threshold where precision equals recall.

Balanced accuracy is used for evaluation in Task 2 to address the imbalance between positive and negative samples in the dataset, where the ratio of negative to positive samples is approximately 10:1 due to the way the training data was constructed.

*D. Saliency Map Interpretation*

To understand which regions of the input sequence the model focuses on when predicting splicing sites, we employed a saliency map technique calculated based on input gradients [24]. For each input DNA sequence $x \in \mathbb{R}^{L \times C}$, where $L$ is the sequence length and $C$ is the number of input channels, the gradient of the model's prediction corresponds to the a input sequence is computed as:

$$\phi G = \frac{\partial x}{\partial y} \in \mathbb{R}^{L \times C} \quad (7)$$

The saliency score of $S_i$ at each position is computed by taking the sum of absolute values of the gradients across all the channels:

$$S_i = \sum_{j=1}^{c} \left|\frac{\partial y}{\partial x_{i,j}}\right| \quad (8)$$

This gives a one-dimensional saliency vector $x \in \mathbb{R}^L$ that highlights the importance of each position of the sequence in influencing the model's prediction.

To analyze general patterns and visualize it, we averaged the saliency maps across all the splicing sequence in each species with 2500 surrounding bases.

$$\bar{S} = \frac{1}{N}\sum_{k=1}^{N} S^{(k)} \quad (9)$$

Where $N$ is the number of splicing sequences of each species, and $S^{(k)}$ is the saliency map of the $k$-th sequence. We then visualized the averaged saliency map $\bar{S}$ within a window of ±2500 bases centered on each splicing site, showing patterns for each species.

In addition to the global saliency profiles around ±2500 bp of the splicing site, we further explored the base-level (A, C, G, T level) importance by constructing Saliency-Weighted Sequence Logos within a narrower window of ±50 bp. For each sequence, we extracted the saliency values and aligned them with the corresponding nucleotide bases in the window. The saliency values were summed per base type (A, C, G, T) and normalized across positions to reflect the relative importance of each base at each position. This visualization method combines gradient-based saliency techniques with sequence logo representations [25] and was implemented using logomaker tool in python environment [26]. The resulting logos helped identify sequence motifs or patterns, such as polyA and poly T stretches, what pattern the model heavily relies on near splicing junctions. Species-wise averaged logos and a combined logo across all species revealed conserved and species-specific importance patterns learned by the model, making it suitable for visualization and interpretation.

## III. EXPRIMENTS

### A. Splicing Site Detection Evaluation

First, CircFormerMoE was pretrained on the splicing site detection task using data from all 10 plant species. During this pretraining, it is compared against the baseline CircFormer model, which uses the same architecture but without the species-specific Mixture-of-Experts (MoE) heads for different species. As illustrated in Fig. 3, after training on the training sets for 500 epochs with the same experimental setting, CircFormerMoE achieved a 4.1% higher mean average recall (mAR) and a 2.3% higher mean average precision (mAP) across all species compared to the CircFormer baseline model, demonstrating the effectiveness of incorporating species-specific MoE heads for capturing diverse splicing patterns while leveraging shared knowledge from the same backbone learned on different species in each batch.

The sharp decline in recall curve during early training is attributed to a 285× weighting applied to the loss of positive positions during the first 10 epochs, which was later removed. It is necessary cause without this initial weighting, the model would fail to learn from the training data, where non-splicing site positions (negatives) vastly outnumber splicing site positions (positives). In other words, since splicing sites constitute only a small fraction of the entire genome, the model would otherwise converge to predicting all positions as non-splicing sites in order to minimize the loss. In addition, balanced sampling was applied during training, ensuring that each batch contained an equal number of samples from each species to promote uniform learning across species.

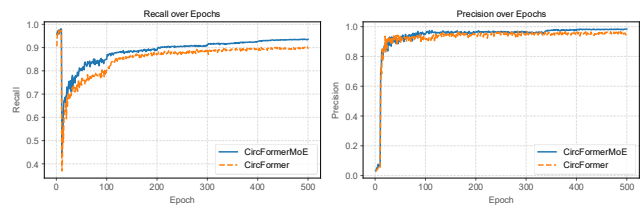

**Fig. 3.** Performance comparison between CircFormerMoE and CircFormer trained across all species. The left panel shows the mean average recall across the 10 species in the evaluation set, while the right panel presents the mean average precision. The values are averaged from diverse results of 10 species.

TABLE II
SSD EVALUATION METRICS AVERAGED ACROSS
EVALUATION SETS OF 10 SPECIES.

| Model | AP (%) | AR (%) | F1 (%) | Top-k acc (%) |
|---|---|---|---|---|
| CircFormer | 0.976 | 0.966 | 0.972 | 0.896 |
| CircFormerMoE | **0.992** | **0.992** | **0.992** | **0.919** |
| SpliceAI | 0.917 | 0.910 | 0.913 | 0.855 |

Next, we conducted transfer learning using pretrained models, fine-tuning them on 10 specific plant species. Three models including CircFormer, CircFormerMoE and SpliceAI were compared. As shown in Table II, CircFormer baseline model outperforms SpliceAI by 5.9% in mean average precision (mAP), 5.6% in mean average recall (mAR), 5.9% in f1-score and 4.1% in top-k accuracy. CircFormerMoE achieves the best performance across all metrics, outperforming SpliceAI by 7.5% in AP, 8.2% in AR, 7.9% in f1-score and 6.4% in top-k accuracy. These results highlight the effectiveness of both the overall model architecture design in capturing circRNA splicing features, as well as the species-specific Mixture-of-Experts (MoE) design in modeling species-specific patterns while leveraging shared knowledge across species.

The detailed evaluation results of the models including CircFormerMoE, CircFormer, and SpliceAI across 10 different species are illustrated in Fig. 4. The results demonstrate that CircFormerMoE consistently achieves superior performance across all evaluated species, outperforming the other two models in nearly every metric. Specifically, CircFormerMoE maintains precision and recall rates above 99% for all ten species, showcasing its robustness in accurately identifying circular RNA splice sites regardless of inter-species genomic variation. Furthermore, the top-k accuracy of CircFormerMoE exceeds 90% on nine out of ten species, with Pisum sativum being the only exception. Notably, all three models show relatively lower performance on the Pisum sativum dataset, which may be attributed to a higher noise level or label inaccuracies in this specific dataset. Except for the recall and top-k accuracy on brassica juncea, the baseline CircFormer model still outperforms SpliceAI, further validating the strength of the proposed architecture.

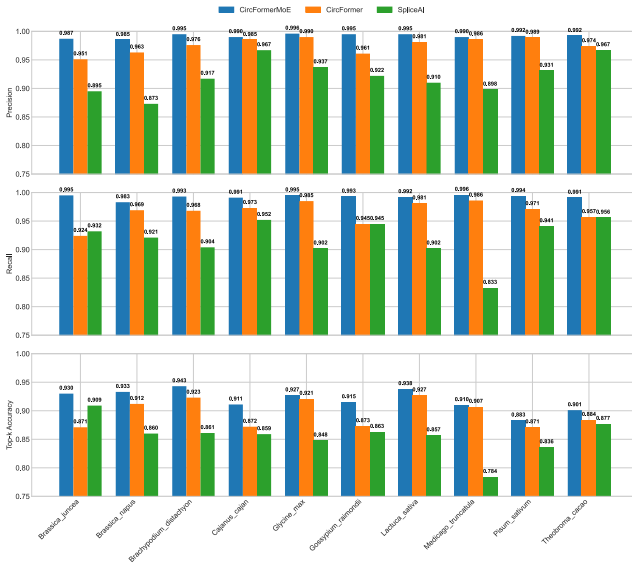

**Fig. 4.** Detailed evaluation results on Task 1 across 10 species. Each subplot shows a comparison among three models including CircFormerMoE, CircFormer and SpliceAI, on one of the following metrics: precision (top), recall (middle), and top-k accuracy (bottom).

*B. Splicing Site Pairing Evaluation*

TABLE III
SSP EVALUATION METRICS OF 10 SPECIES

| Species | Acc (%) | P (%) | R (%) | F1 (%) |
|---|---|---|---|---|
| Brassica juncea | 95.1 | 92.4 | 88.1 | 90.2 |
| Brassica napus | 96.3 | 96.9 | 91.2 | 94.0 |
| B. distachyon | 97.6 | 97.8 | 92.3 | 95.0 |
| Cajanus cajan | 98.5 | 97.3 | 90.5 | 93.8 |
| Glycine max | 99.0 | 98.5 | 92.1 | 95.2 |
| Gossypium raimondii | 96.1 | 94.5 | 98.3 | 96.4 |
| Lactuca sativa | 100.0 | 100.0 | 100.0 | 100.0 |
| Medicago truncatula | 98.6 | 98.6 | 90.7 | 94.5 |
| Pisum sativum | 98.9 | 95.1 | 91.1 | 93.1 |
| Theobroma cacao | 97.4 | 95.7 | 88.4 | 91.9 |
| TOTAL | 97.35 | 96.68 | 92.27 | 94.41 |

In Task 2, where model focuses on circRNA splicing site pairing and classification, the model was initialized with the backbone pretrained on Task 1 (circRNA splicing site detection). Positive samples in the dataset were constructed by concatenating two splicing sites along with their surrounding ±500 bp sequences, with a spacer token "MMMMM" inserted at the junction. The ±500 bp context was empirically selected as a trade-off between classification accuracy and inference efficiency, as extending the sequence length beyond this window size provided negligible performance gains but obviously adds more computational cost. Negative samples were constructed by samely concatenating splicing site pairs but originating from different circRNAs. This allows the generation of a large number of negative examples. To maintain a balanced training set, we sampled a set of negative examples at a ratio of approximately 10:1 relative to the number of positive samples. And similar to Task 1, We assign weights to the loss of a few sample categories in the early stages of training to enhance the stability of the training

Trained on splicing site pairing, CircFormerMoE achieves an accuracy of 97.35%, which is averaged on positive class and negative class, a precision of 96.68%, a recall of 92.27% and a F1-score of 94.41%. The performance is summarized in Table III. To better visualize the per-species classification performance, Fig. 5 presents a heatmap of the four metrics. The model consistently achieves average accuracy above 90% and f1-scores above 90%, demonstrating high accuracy and a balanced precision-recall trade-off.

Notably, Lactuca sativa reaches perfect scores (100%) across all four metrics, indicating extremely high confidence and correctness in its predictions. This may be attributed to cleaner annotations or more distinct sequence patterns in this species. Other species such as Glycine max, Pisum sativum, and Cajanus cajan also exhibit near-perfect performance, suggesting the model's strong generalization to capture diverse splicing signals.

Conversely, Brassica juncea and Theobroma cacao show slightly lower recall and F1-scores compared to other species (e.g., 88.1% recall in Brassica juncea), possibly due to noisier data, higher sequence variability, or label ambiguity. These findings highlight potential areas for dataset refinement or tailored fine-tuning.

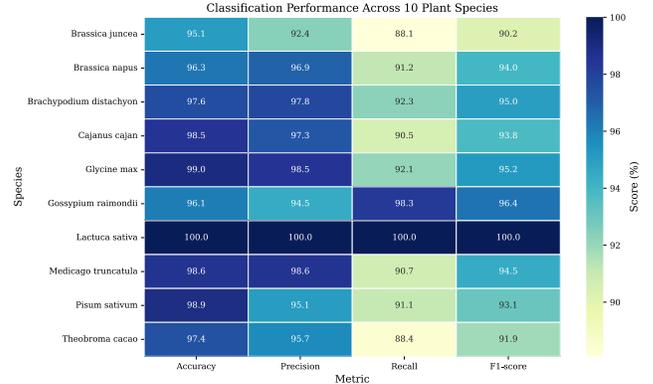

**Fig. 5.** Heatmap of evaluation metrics on Task 2 across 10 species. Each row corresponds to a species and each column to a metric on the test set. The color intensity reflects the performance score, with higher values indicating better performance.

*C. Model Interpretability*

To understand which regions of the input sequence the model emphasizes when detecting splicing sites, we computed saliency maps for all circRNA splicing site sequence across 10 species, see Fig. 6. For each splicing site, the saliency map was calculated based on its surrounding bases. These saliency maps were then averaged across all splicing site sequences within each species to produce a species-level importance profile. While multiple peaks appear across the positions, the regions immediately surrounding the splicing site exhibit the highest importance, indicating that the model focuses more on these narrow positions for making predictions.

Based on positional importance, we then focused on a narrower window of ±50 base pairs (bp) around the splice sites to analyze base-level contributions in greater detail. For each sequence within this window, we extracted the saliency values corresponding to each nucleotide (A, C, G, T) and aggregated these values across all sequences and all species. We then normalized the aggregated saliency scores at each position to obtain a relative importance distribution for each base. This enabled us to generate Saliency-Weighted Sequence Logos that visually represent the model's preference for specific nucleotides at precise positions near the splice junction. Fig. 7 is the visualization of Saliency-Weighted Sequence Logo, it reveals that continuous stretches of adenine (A) and thymine (T) nucleotides exhibit notably high importance, indicating their significant role in the model's predictions.

centered on the splice site, averaged across all species. Each position shows the normalized contribution of adenine (A), cytosine (C), guanine (G), and thymine (T) to the model's splicing site prediction. Notably, continuous stretches of adenine (A) and thymine (T) display higher importance, indicating their important role in the model's decision-making process.

Based on this visualization, we scanned the circRNA sequences in our dataset to identify the presence of obvious polyA or polyT stretches (Continuous ≥ 5 bases) within a 100-base window around the splice site. Compared to theoretical random sequences, where approximately 17.4% of sequences contain such polyA/T regions, our real data showed a markedly higher occurrence. Specifically, 43.7% of sequences in our dataset contained polyA/T stretches within this region. This significant enrichment suggests that polyA/T sequences are biologically relevant features that may contribute importantly to the model's splice site predictions of circRNAs.

## IV. DISCUSSION

In this study, we propose a lightweight framework called CircFormerMoE for directly predicting circular RNA from plant genomic DNA sequences. We began by asking two questions: How can we accurately predict ultra-long circular RNAs from genomes with each chromosme spanning tens of millions of bases? And how can we design a model that is as lightweight as possible while keeping high performance? To address these, our framework decomposes the problem into two subtasks: splicing site detection (SSD, Task 1) and splicing site pairing (SSP, Task 2). By focusing on these subtasks, the model processes only relevant sequence segments rather than entire genomes, making prediction of ultra-long circular RNAs computationally feasible. The deep learning architecture incorporates lightweight modules optimized for efficient inference. for example, on widely available GPUs for personal use like the NVIDIA RTX 4080, it can process over one million bases per second during inference.

Although Task 2 is relatively simpler and could benefit from a faster model to accelerate inference, our current implementation still encounters some bottlenecks in speed, Furthermore, it could be better addressed by leveraging broader and more datasets to improve its robustness and accuracy. Since this task is a classical problem in deep learning for genomes, models that perform well on human genes can often be applied to plants as well. For example, Splam [27], which was developed to predict splice junctions in DNA using deep residual convolutional neural networks, can still be effective for pairing splicing sites of circular RNA.

Initially, we also explored the inclusion of human genomic data for circRNA prediction. Interestingly, while both the recognition of a position as a donor or acceptor and the pairing of donor and acceptor splice sites to form circRNA can be effectively modeled and classified using deep learning models, individual splice sites in human genomes could not be learned and detected using deep learning models that worked well for plant genomes. This observation suggests that the signal indicating whether two splice sites will pair to form a circRNA may lie not in their individual features, but in their joint context and interaction. Instead, It is likely that this signal arises from their combined sequence context or other properties that only become evident when the two sites are considered together. In the future, it will be important to design models that can better capture such dependencies, possibly by integrating structural context, or regulatory features. Understanding these differences could also help to distinguish mechanisms of circRNA biogenesis in human and other species.

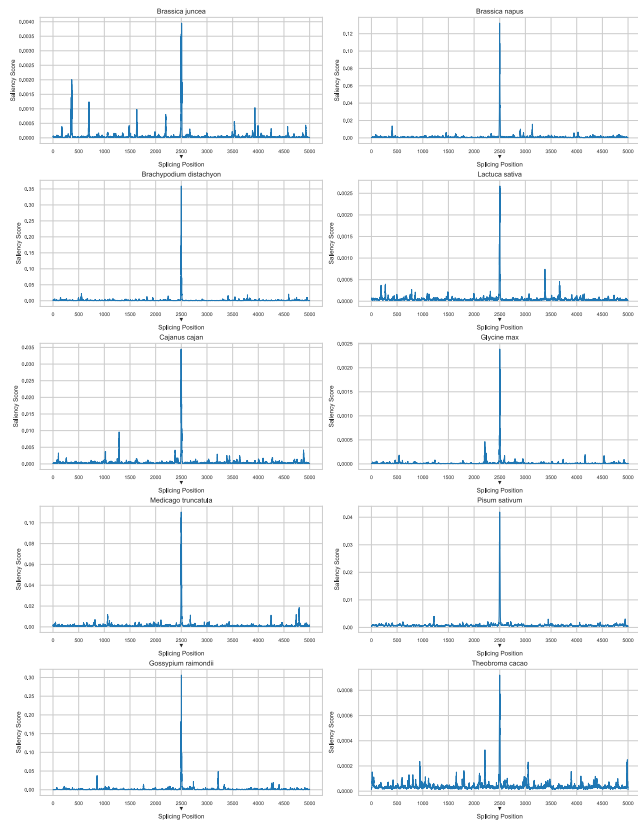

**Fig. 6.** Saliency map of position importance of the model prediction surrounding circularRNA splicing sites, where the middle position corresponds to the splice sites.

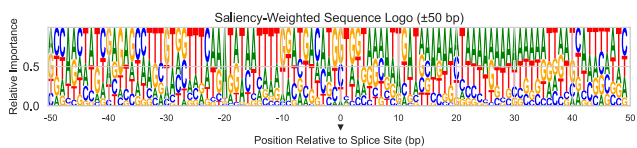

**Fig. 7.** Saliency-Weighted Sequence Logo represents the relative importance of nucleotides within a ±50 bp window